\documentclass[preprint,5p, times, twocolumn]{elsarticle}

\usepackage{amssymb}
\usepackage{amsmath}

\usepackage{subcaption}
\usepackage{subfig}
\usepackage{svg}
\usepackage{booktabs}
\usepackage{multirow}
\usepackage{booktabs}
\usepackage{xcolor}
\usepackage{natbib}

\journal{Information Sciences}

\begin{document}

\begin{frontmatter}



\title{On Preserving Geometrical Invariance for Superpixel Image Classification using Graph Transformer}




\author[1]{Sarabeshwar Balaji\corref{cor1}}
\ead{sarabeshwar23@iiserb.ac.in}

\author[1]{Shubham Mohanty}
\author[1]{Akash Anil}

\affiliation[1]{
    organization={Indian Institute of Science Education and Research Bhopal},
    city={Bhopal},
    state={Madhya Pradesh},
    postcode={462066},
    country={India}
}

\cortext[cor1]{Corresponding author}

\begin{abstract}
Convolutional Neural Network (CNN) and Vision Transformer (ViT) for image classification exploit a dense grid of pixels containing redundant information. Consequently, for a larger image dataset, CNNs and ViTs face deployability challenges due to high computational complexity. Representing images as graphs of superpixels offers an efficient alternative that preserves key information while eliminating pixel-level redundancy. Graph Neural Networks (GNNs) have been utilized on such graphs to perform image classification. However, GNNs are known to struggle with capturing long-range dependencies which is important in the domain of image classification. Furthermore, a majority of these superpixel-based image classification approaches do not explicitly preserve translation/rotation invariance. Nevertheless, preserving translation/rotation invariance is important for robust image classification. Thus, this paper proposes \texttt{SuperGT}, a Graph Transformer-based framework for image classification, which captures the long range dependencies, along with a pre-processing scheme that preserves translation/rotation invariance. We evaluate \texttt{SuperGT} on CIFAR-10 dataset and observe that it performs significantly better than many baselines. Furthermore, we note that the overall performance of \texttt{SuperGT} is comparable to the previous state-of-the-art model, namely, \texttt{ShapeGNN}, without relying on coordinates of the boundary points of each superpixel required by \texttt{ShapeGNN}.

\end{abstract}




\end{frontmatter}



\section{Introduction}
\label{sec:intro}

Deep learning models such as Convolutional Neural Networks (CNNs) \cite{simonyan2014very} and Vision Transformer(ViTs) \cite{dosovitskiy2020image} have shown remarkable capability in image classification performance and have been adopted across several fields, including geophysics \cite{yu2021deep}, bioinformatics \cite{min2017deep} and remote sensing \cite{maggiori2016convolutional}. These deep learning models mainly operate on dense grid-based representations of images, requiring computation over large regions that may contain limited task-relevant information. In particular, CNNs often struggle to capture long range dependencies because of their local receptive fields. Consequently, there is a requirement to increase depth of the network. Although ViTs propose solutions for capturing long range dependencies, they are challenged with quadratic computational requirements due to the self-attention mechanism \cite{vaswani2017attention}.

To alleviate such computational challenges associated with images, researchers have explored grouping structurally correlated pixels into compact regions known as superpixels~\cite{cosma2023geometric, long2021graph, avelar2020superpixel, bae2022superpixel, youn2021dynamic}. A superpixel is defined as a group of adjacent pixels demonstrating similar characteristics such as color and intensity. An image can be represented as a graph with superpixels as nodes while edges are constructed by heuristics such as $k$-nearest neighbors or region adjacency.~\cite{rodrigues2024graph}. Such a representation preserves key characteristics while reducing the pixel-level redundancy. Such graphs open a new direction in image classification capable of harnessing the popular graph-based models such as Graph Neural Network (GNN)~\cite{avelar2020superpixel, cosma2023geometric, bae2022superpixel, long2021graph, youn2021dynamic}. Although GNNs have shown impressive capabilities in solving many problems for graph-based datasets, they are inherently limited in capturing long-range dependencies due to over-smoothing \cite{oonograph} and over-squashing \cite{alon2021bottleneck, topping2021understanding}. However, for image classification, the superpixel graphs may possess semantically important superpixels far apart and may require modeling the long-range dependencies. Consequently, many recent graph learning methods have adopted Graph Transformers \cite{dwivedi2020generalization, kreuzer2021rethinking, rampavsek2022recipe, wu2021representing}, which leverage the attention mechanism to allow each node to attend to all other nodes in the graph. Although GNN frameworks exploiting Graph Transformers seem to capture structural characteristics as well as global relatedness, to the best of our knowledge, they have not been used for superpixel-based image classification. 

Graph Transformers (GTs) are one form of GNNs designed to capture long-range dependencies while maintaining permutation equivariance at the node level and permutation invariance in graph level representations~\cite{dwivedi2020generalization}. However, GTs and GNNs do not inherently preserve translation or rotation invariance in graph level representations, which are important characteristics to be preserved for image classification. To the best of our knowledge, existing GNN-based frameworks for superpixel-based image classification do not explicitly preserve translation and rotation invariance~\cite{avelar2020superpixel, cosma2023geometric, bae2022superpixel, long2021graph, youn2021dynamic}. Thus, we arrive at the following research question: \textit{how can GTs be utilized on superpixel graphs for image classification while preserving translation and rotation invariance?}

To answer the above question, this paper attempts to incorporate translation invariance with GTs using mean centering of the spatial node features. Moreover, we utilize a Principal Component Analysis (PCA) \cite{hotelling1933analysis}-based approach which is capable of preserving translation and rotation invariant simultaneously. We evaluate the proposed mechanism that preserves translation and rotation invariance within a Graph Transformer framework for the image classification using superpixel graphs constructed from CIFAR-10 dataset~\cite{krizhevsky2014cifar}. We refer to the proposed approach as \texttt{SuperGT} \footnote{Our code is available at: https://github.com/SarabeshwarBalaji/SuperGT} in the subsequent parts of the paper. We compare the performance of \texttt{SuperGT} to the state-of-the-art GNN-based models. We note that the \texttt{SuperGT} yields comparable results with respect to the state-of-the-art model, namely, \texttt{ShapeGNN}~\cite{cosma2023geometric}. However, for the experiments considering varying percentages of available training samples, we observe a lower and unstable performance by \texttt{ShapeGNN} in comparison to \texttt{SuperGT} and other GNN-based models. 

In summary, this paper has the following contributions:

\begin{enumerate}
    \item We propose a framework \texttt{SuperGT} for image classification using Graph Transformers capable of capturing global relatedness along with preserving translation and rotation invariance.  

    \item We propose to utilize Principal Component-based pre-processing scheme to preserve both translation and rotation invariance simultaneously and can be applied to any superpixel-based image classification frameworks.
    
    \item We conduct extensive experiments demonstrating the effectiveness of \texttt{SuperGT} against standard GNN-based methods and  state-of-the-art superpixel based image classification frameworks. Furthermore, we provide critical analysis to the observed performances using the lens of variations in availability of the training samples. 
\end{enumerate}

The remainder of the paper is organized as follows: Section~\ref{sec:related_works} discusses the related works in the direction of image classification using superpixel-based graphs and GNNs. Section~\ref{sec:methods} discusses the image to graph transformation procedure, the PCA-based pre-processing scheme, and the GT backbone. Section~\ref{sec:experiments} discusses the best positional and structural augmentation of graphs, the efficacy of \texttt{SuperGT} compared to popular GNN and state-of-the-art models in the literature, and the ablation study of \texttt{SuperGT}. 

\section{Related Works}
\label{sec:related_works}

This section discusses some of the related works that exploit superpixel graphs for image classification using graph-based machine learning. In particular, we focus mainly on studies exploiting graph neural networks as recently, such methods have shown remarkable performance for image classification~\cite{avelar2020superpixel, cosma2023geometric, bae2022superpixel, long2021graph, youn2021dynamic}.

To construct superpixel graphs, most studies follow two paradigms for edge construction for a given set of superpixel nodes: (i) the edges are constructed between two nodes if their corresponding superpixels are adjacent to each other in the image, commonly referred to as a Region Adjacency Graph (RAG)~\cite{cosma2023geometric, avelar2020superpixel}; or (ii) the edges are constructed between two nodes based on node feature similarity often measured by K-Nearest Neighbors (KNN)~\cite{youn2021dynamic, bae2022superpixel, long2021graph}. Using the first paradigm of superpixel graph construction, Avelar et al.~\cite{avelar2020superpixel} proposed utilizing Graph Attention Networks~\cite{velivckovic2018graph} on RAG of the image for classification. We refer to this as \texttt{RAG-GAT} in our work. Long et al.~\cite{long2021graph} proposed \texttt{HGNN}, a framework uses the KNN approach for graph construction and showed an improvement in image classification by overcoming over-smoothing issues by using residual connections. Linh and Youn et al.~\cite{youn2021dynamic} proposed \texttt{DISCO-GCN} that treats superpixels as point clouds and dynamically reconstructs graphs after every convolution layer using KNN technique. Bae et al.~\cite{bae2022superpixel} proposed \texttt{IMGCN-LPE} framework that initializes node positions using random walks and updates them during the training cycle for addressing the limitations of fixed positional encodings. 

Recently, Cosma et al.~\cite{cosma2023geometric} proposed \texttt{ShapeGNN} which evaluates with both paradigms of graph construction and empirically achieved a better result with RAG-based method. \texttt{ShapeGNN} incorporated superpixel shape information into the feature vector and produced shape encoding using GNNs over superpixel boundary. Presently, to the best of our knowledge \texttt{ShapeGNN} is the state-of-the-art method proposed for superpixel-based image classification.

All the above studies focused on improving the classification performance by exploiting a graph construction paradigm along with an efficient GNN-based model. However, this paper focuses on understanding basic limitations with GNNs exploiting superpixel-based graphs for image classification task. Moreover, this paper attempts to revisit the preservation of fundamental properties such as translation and rotation invariance critical in image classification task.

\section{Methodology}
\label{sec:methods}
This section discusses the methodologies along with key techniques used for the proposed framework \texttt{SuperGT}. Firstly, we revisit the approach used to build the superpixel graph which is followed by proposed methodologies to preserve the translation and rotation invariance for superpixel-based image classification. Finally, we discuss the GT backbone used in \texttt{SuperGT}.

\subsection{Image to Graph Transformation}
\label{sec:image_to_graph}
\begin{figure*}[t]
    \centering
    \begin{subfigure}[b]{0.32\linewidth}
        \includegraphics[width=\linewidth]{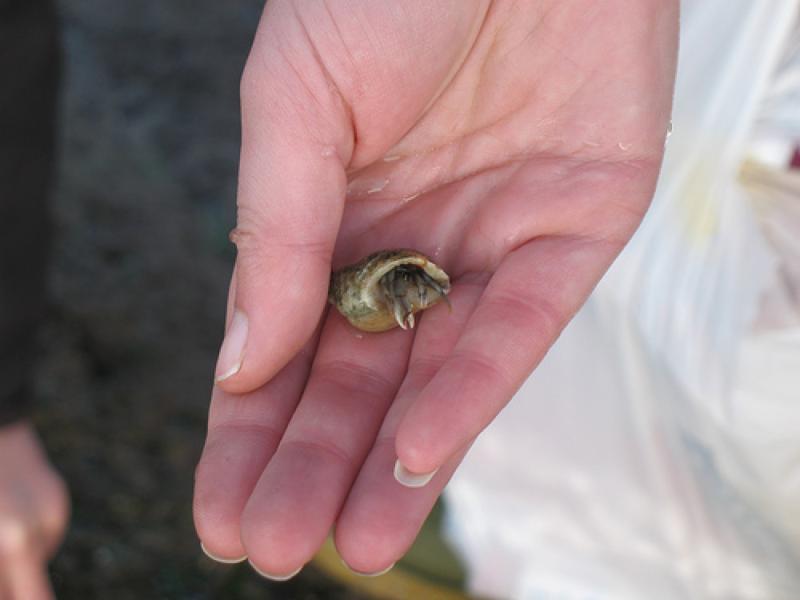}
        \caption{}
        \label{fig:input}
    \end{subfigure}
    \hfill
    \begin{subfigure}[b]{0.32\linewidth}
        \includegraphics[width=\linewidth]{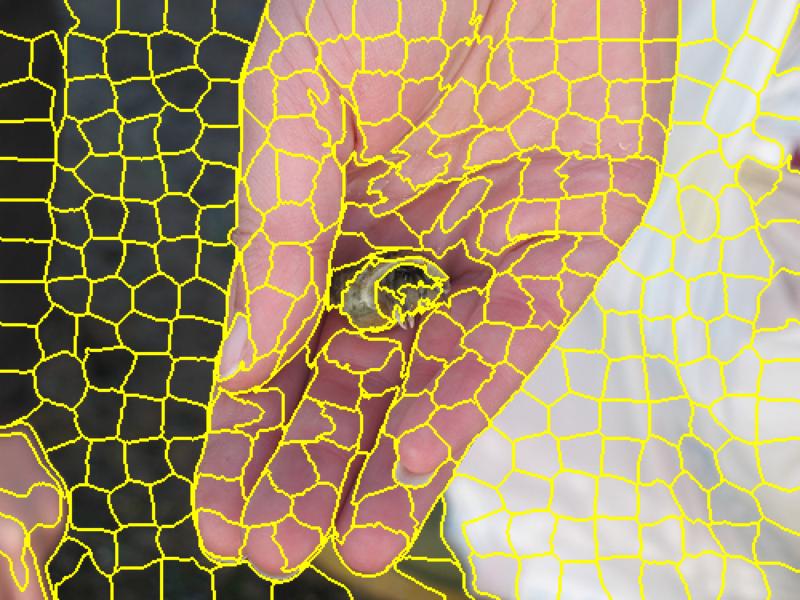}
        \caption{}
        \label{fig:slic}
    \end{subfigure}
    \hfill
    \begin{subfigure}[b]{0.32\linewidth}
        \includegraphics[width=\linewidth]{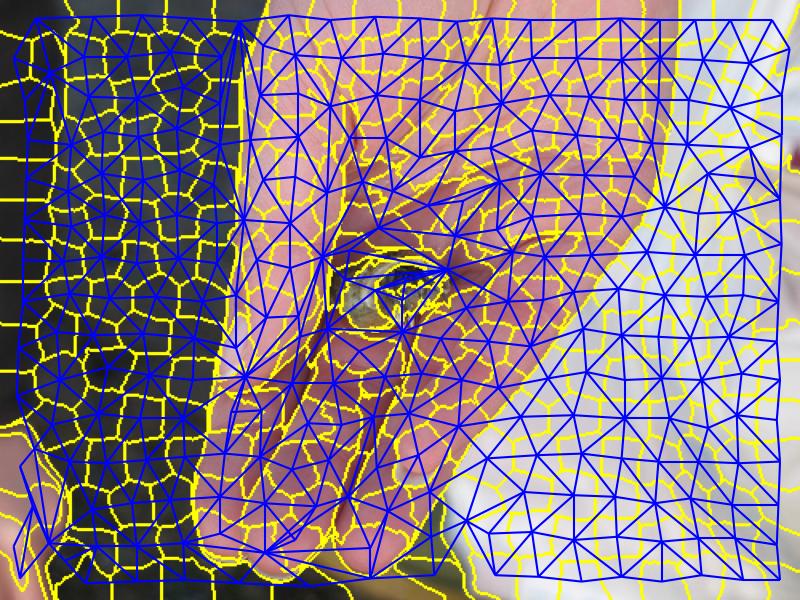}
        \caption{}
        \label{fig:rag}
    \end{subfigure}
    \caption{\textbf{Illustration of the superpixel graph construction} (a) Original input image (from ImageNet \cite{deng2009imagenet}). (b) SLIC segmentation boundaries overlaid on the image. (c) Region Adjacency Graph (RAG), where blue line segments indicate edges between neighboring superpixels.}
    \label{fig:slic_rag_pipeline}
\end{figure*}

The aim of superpixel segmentation algorithms is to cluster pixels of a given image that share various properties such as color and spatial location. There are many algorithms for segmenting images into superpixels. Among these methods, Simple Linear Iterative Clustering (SLIC)\cite{achanta2012slic} is the most widely used. We utilize SLIC in our framework for generating the main results. The resulting superpixel graph encodes node's attribute/feature in terms of mean color, color variance, fraction of region covered by the superpixel in the image, and cartesian coordinates of the centroid of the superpixel. Therefore, the node feature vector has a total of nine dimensions. 
All node features are normalized to the range [0,1]. As discussed in Section~\ref{sec:related_works}, a superpixel-based image representation can be constructed as RAG or KNN graph. We employ RAG approach as demonstrated in the Figure~\ref{fig:slic_rag_pipeline}. 

\subsection{Geometric Invariance} Ideally, for image classification, the framework should be invariant to the rotations and translations of the underlying image. To the best of our knowledge, none of the superpixel-based image classification frameworks explicitly provide either translation or rotation invariance. Recently, a notable attempt was made by Cosma et al.~\cite{cosma2023geometric}, who explored geometric invariance through the proposed \texttt{ShapeGNN} framework. In particular, they considered replacing the GNN backbone of \texttt{ShapeGNN} with an \texttt{EGNN} \cite{satorras2021n} to incorporate geometrical equivariance. Further, they also investigated simple pre-processing strategies to introduce rotation invariance. However, these approaches resulted in reduction of the classification performance. These observations highlight the challenges of incorporating invariance into superpixel-based image classification frameworks while maintaining classification performance.

\subsubsection{Preserving Geometrical Invariance using Principal Component Analysis (PCA)~\cite{hotelling1933analysis}}
\label{subsec:pca}
For superpixel-based image classification preserving geometrical invariance, we propose to utilize a PCA-based pre-processing scheme that can incorporate both translation and rotation invariance. Let $G_i$ be a graph that corresponds to the image $I_i$. Let $F_i$ denote the node feature matrix of $G_i$ where $F_i \in \mathbb{R}^{n \times 9}$, and $n$ denotes 
the number of nodes in the graph. Let $C_i \in \mathbb{R}^{n \times 2}$ be the 
submatrix of $F_i$ which carries the centroid  coordinates of the superpixels. Presuming the image segmentation algorithm is translation and rotation invariant, such 
actions on $I_i$ would only transform the $C_i$ matrix while remaining node attributes are unchanged.

Principal Component Analysis (PCA) is primarily used for dimensionality reduction in machine learning. It can be shown that the PCA transformation of a given data point is invariant to rotation and translation of the input set of data points. Several approaches leveraging this property have been proposed in point cloud analysis~\cite{bezick2025robust, xiao2020endowing}. We apply PCA with 2 principal components to the matrix $C_i$, considering each centroid row as an input data point. The mean centroid is defined as $\mu = \frac{1}{n}\sum_{i=1}^{n} c_i$, where $c_i \in \mathbb{R}^2$ 
denotes the $i$-th row of $C_i$. The canonical representation for a centroid $z_i$ is given by:
\begin{equation}
    z_i = \begin{pmatrix} v_1^\top (c_i - \mu) \\ v_2^\top (c_i - \mu) \end{pmatrix}
    \label{eq:1}
\end{equation}
where $v_1, v_2 \in \mathbb{R}^2$ are the first and second principal components 
(eigenvectors of the covariance matrix $\Sigma = \frac{1}{n}\tilde{C}_i^\top 
\tilde{C}_i$, with $\tilde{C}_i = C_i - \mathbf{1}_n\mu^\top$), ordered by descending 
eigenvalue. We will now formally show that PCA is capable of preserving both translation and rotation invariance properties.

\paragraph{\textbf{Translation Invariance}} Here, let $z'_i$ denote the canonical representation obtained after applying a translation to the image. Let each centroid be translated by $t \in \mathbb{R}^2$, i.e.\ $c' = c + t$. The new 
mean will be $\mu' = \mu + t$. Substituting these in Equation~\ref{eq:1}:
\begin{equation}
    z'_i = \begin{pmatrix} v_1^\top (c'_i - \mu') \\ v_2^\top (c'_i - \mu') \end{pmatrix}
       = \begin{pmatrix} v_1^\top (c_i + t - \mu - t) \\ 
         v_2^\top (c_i + t - \mu - t) \end{pmatrix}
       = \begin{pmatrix} v_1^\top (c_i - \mu) \\ v_2^\top (c_i - \mu) \end{pmatrix} = z_i
\end{equation}
Therefore, $z'_i = z_i$ for all centroids, establishing the translation invariance of the canonical representation.

\paragraph{\textbf{Rotation Invariance}} Here, let $z'_i$ denote the canonical representation obtained after applying a rotation to the image. Let each centroid be rotated by a $2{\times}2$ rotation matrix $R$, where $RR^\top = I$. The transformed centroid is given by $c'_i = Rc_i$, and the corresponding mean becomes $\mu' = R\mu$. The centered coordinates transform as $\tilde{C}_i' = \tilde{C}_iR^\top$,and therefore the covariance matrix becomes $\Sigma'=\frac{1}{n}\tilde{C}_i'^\top\tilde{C}_i'=R\Sigma R^\top$. Let $v_k$ be an eigenvector of $\Sigma$ with eigenvalue $\lambda_k$. Then $\Sigma'(Rv_k)=R\Sigma R^\top Rv_k=R\Sigma v_k=\lambda_k(Rv_k)$, showing that $Rv_k$ is an eigenvector of $\Sigma'$ with the same eigenvalue. Hence, the principal components rotate together with the data, i.e., $v_k' = Rv_k$ for $k=1,2$. We resolve this sign ambiguity of eigenvectors by checking the sign of its $y$-component and replacing $v_k$ with $-v_k$ if $v_{k,y} < 0$, and leaving as it is otherwise.

Substituting into Equation~\ref{eq:1},
\begin{equation}
    z'_i
    =
    \begin{pmatrix}
        v_1'^\top (c'_i-\mu')\\
        v_2'^\top (c'_i-\mu')
    \end{pmatrix}
    =
    \begin{pmatrix}
        (Rv_1)^\top(Rc_i-R\mu)\\
        (Rv_2)^\top(Rc_i-R\mu)
    \end{pmatrix}
    =
    \begin{pmatrix}
        v_1^\top(c_i-\mu)\\
        v_2^\top(c_i-\mu)
    \end{pmatrix}
    =
    z_i.
\end{equation}
Therefore, $z'_i = z_i$ for all $i$, establishing the rotation invariance of the canonical representation.

In essence, mean-centering is responsible for translation invariance, and as the principal components are equivariant with respect to rotations of the input set of data points, resolving a data point with respect to them produces a rotation invariant representation.

\paragraph{\textbf{Empirical Remarks}} During empirical evaluation, we found that PCA-based scheme preserving translation and rotation invariance simultaneously resulted with a slight decrease in accuracy for image classification. Thus, in this paper, all the subsequent experiments employ only mean-centering of centroid coordinates, thereby preserving only the translation invariance. 

\begin{figure*}[htbp]
    \centering
    \includegraphics[width=\linewidth]{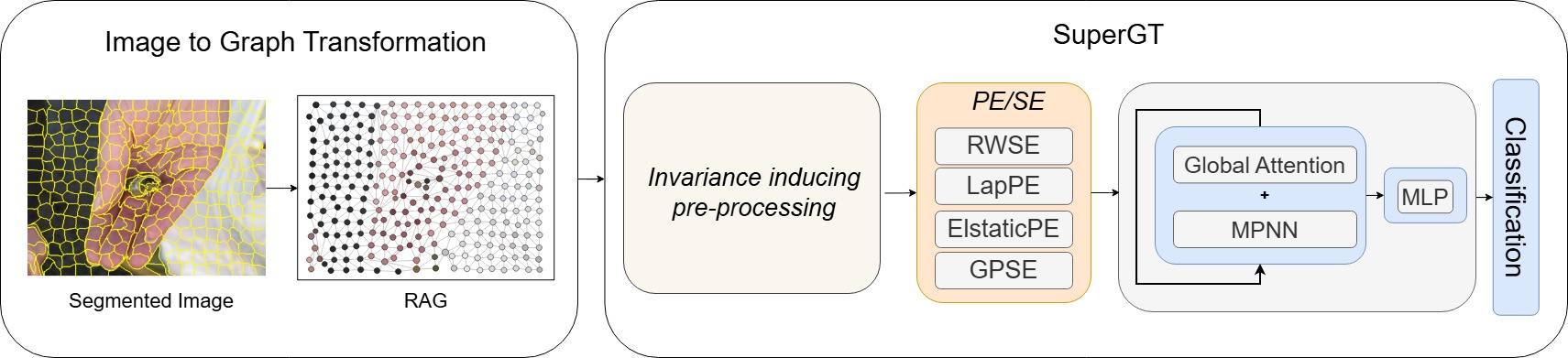}
    \caption{\textbf{Proposed framework.} Each image is first segmented into superpixels, from which a Region Adjacency Graph (RAG) is constructed. For visualization, here the graph nodes are colored according to the mean color of their corresponding superpixels. We then apply the proposed feature preprocessing scheme to incorporate geometric invariance, including translation and rotation invariance. Then positional and structural encoding (PE/SE) is concatenated with node features before being processed by a graph Transformer architecture consisting of message-passing and global attention layers. Several such layers are used and the resulting node embeddings are pooled and passed to a multilayer perceptron (MLP) for final image classification.}
    \label{fig:supergt}
\end{figure*}

\subsection{Graph Transformers}
\label{subsec:gt}
Motivated with the capabilities and research gaps with recently proposed GNNs for superpixel-based image classification, this paper uses Graph Transformer (GT), a GNN replacing traditional message passing framework with self-attention mechanism capturing local and global dependencies between relevant nodes in the graph~\cite{dwivedi2020generalization, kreuzer2021rethinking, rampavsek2022recipe, wu2021representing}. GTs are capable of harnessing both structural and positional information and thus are more expressive at capturing long-range dependencies.  


\paragraph{\textbf{Positional and Structural Encoding}} GTs usually treat the input graph as a fully connected graph. This leads to loss of structural information (adjacency) that is inherent with the graph \cite{battaglia2018relational}. For this purpose, positional encoding (PE) and structural encoding (SE) are concatenated with the original node features so that the structural biases can be reintroduced. PE/SE aids in identifying the structural and semantic properties of nodes in the graph and the subgraph respectively \cite{dwivedi2020generalization,rampavsek2022recipe}. 
Although there are abundant number of PEs have been proposed in past, we evaluate a commonly used PE based on eigenvectors of the Laplacian matrix (LapPE) \cite{kreuzer2021rethinking, dwivedi2023benchmarking}. Moreover, we also evaluate a Laplacian-based positional encoding (ElstaticPE) derived from electrostatic interactions between nodes \cite{kreuzer2021rethinking, canturkgraph}. For SE, a commonly used approach is the diagonal of the $m$-step random walk matrix (RWSE) \cite{dwivedigraph}. We explore such traditional PE/SE with the superpixel graph along with representations from \textit{Graph Positional and Structural Encoder} (GPSE) \cite{canturkgraph}, a pre-trained foundational model for producing rich structurally informed node embeddings. 

\paragraph{\textbf{Architecture:}} In recent past, several architectures of GTs have been proposed \cite{chen2022structure, hussain2022global, park2022grpe, dwivedi2020generalization, wu2021representing, rampavsek2022recipe}. This paper uses the GT framework proposed by Ramp{\'a}{\v{s}}ek et al. (GraphGPS)\cite{rampavsek2022recipe} to build the graph transformer layer for \texttt{SuperGT} model. A layer in \texttt{SuperGT} can be defined by the following equations:

\begin{align}
    \mathbf{X}_{M}^{\ell+1} &= \text{MPNN}^{\ell}\left(\mathbf{X}^{\ell}, \mathbf{A}\right) \\
    \mathbf{X}_{T}^{\ell+1} &= \text{GlobalAttn}^{\ell}\left(\mathbf{X}^{\ell}\right) \\
    \mathbf{X}^{\ell+1} &= \text{MLP}^{\ell}\left(\mathbf{X}_{M}^{\ell+1} + \mathbf{X}_{T}^{\ell+1}\right)
\end{align}

For $l_{th}$ layer, $\mathbf{X}^{\ell}$ denotes the input nodes features along with the PE/SE. $\mathbf{A}$ denotes the adjacency matrix of the input graph. We utilize GAT \cite{velivckovic2018graph} convolution policy for the $\text{MPNN}^{\ell}$. The $\text{GlobalAttn}^{\ell}$ denotes the global attention that allows the nodes to attend all the other nodes. Our model uses 7 such layers to derive rich node embeddings which are then pooled and passed into a neural network for the final multi-class prediction. We summarize the \texttt{SuperGT} framework in Figure~\ref{fig:supergt}. 


\subsection{Training SuperGT}
We train the model with the pre-defined training split of CIFAR-10~\cite{krizhevsky2014cifar}, which contains 50K images, i.e., 500 images for each class. We evaluate on test split, which contains 10K images. We utilize AdamW~\cite{loshchilovdecoupled} with a weight decay of 7.7e-04. We use the cosine annealing scheduler proposed by Loshchilov et al.~\cite{loshchilov2017sgdr} with initial learning rate as 9.0e-04 and minimum learning rate as 2.5e-05. We use cross entropy loss with a label smoothing factor of 5e-02 to improve generalization over the data with a batch size of 512. The training was conducted for 150 epochs. The hyperparameter search for learning rate, dropout, weight decay and batch size were determined using a Bayesian optimization framework namely, Optuna~\cite{akiba2019optuna}. For this search, 10K images were sampled from the training split while respecting the class distribution, i.e., 100 images for each class.

\section{Experiments}
\label{sec:experiments}
We evaluate the efficacy of \texttt{SuperGT} on CIFAR-10 dataset \cite{krizhevsky2014cifar} against standard GNN architecture and state-of-the-art segmentation-based GNNs for image classification with the aim of answering the following questions:

\begin{itemize}
    \item[\textbf{1.}]Which positional and structural encoding (PE/SE) scheme is the most effective for \texttt{SuperGT}? 
    \item[\textbf{2.}] How does \texttt{SuperGT} compare to existing models in literature for superpixel based graph classification and how does its performance change under limited training data? 
    \item[\textbf{3.}] How robust is \texttt{SuperGT} to noisy data during inference? 
    
\end{itemize}

While the main results uncover the above questions, we also perform some ablation studies considering PCA-based preprocessing, KNN graph construction, and incorporating shape encoding using Hu-moment~\cite{hu2020open}.

\begin{table}[t]
\centering
\caption{\textbf{Classification Accuracy using different PE/SE Augmentations}: 
The red color highlights the best performer, followed by green as second best, and blue signifies the third best.}
\label{tab:pse_comparison}
\resizebox{\columnwidth}{!}{%
\begin{tabular}{llll}
\toprule
\textbf{Category} & \textbf{PE/SE} & \textbf{Transformer} & \textbf{Performer} \\
\midrule
None & - & 78.53 & 79.69 \\
\midrule
\multirow{2}{*}{Structural Encoding}
 & RWSE-16 & 78.70 & 78.91 \\
 & RWSE-20 & 78.43 & 78.44 \\
\midrule
\multirow{3}{*}{Positional Encoding}
 & LapPE-8 & \textcolor{green}{\textbf{78.98}} & 79.58 \\
 & LapPE-10 & \textcolor{red}{\textbf{79.36}} & 79.87 \\
 & ElstaticPE & 76.76 & 78.80 \\
\midrule
\multirow{5}{*}{GPSE}
 & GPSE$_{\text{ZINC}}$    & 78.44 & \textcolor{green}{\textbf{80.09}} \\
 & GPSE$_{\text{GEOM}}$    & \textcolor{blue}{\textbf{78.94}} & \textcolor{red}{\textbf{80.19}} \\
 & GPSE$_{\text{ChemBL}}$  & 78.57 & 79.70 \\
 & GPSE$_{\text{MolPCBA}}$ & 78.74 & \textcolor{blue}{\textbf{79.98}} \\
 & GPSE$_{\text{PCQM4Mv2}}$& 78.87 & 79.96 \\
\bottomrule
\end{tabular}
}
\end{table}

\begin{table}[t]
\centering
\caption{\textbf{Performance of \texttt{SuperGT} compared to the baselines}. We compare our model (\textit{SuperGT}) with standard GNNs and segmentation-based GNNs frameworks in the literature. We report the mean performance across 5 seeds. Results marked with $^\dagger$ are reported directly from the original paper. The red color highlights the best performer,
followed by green as second best, and blue signifies the third best result.}
\label{tab:model_comparison}
\begin{tabular}{l p{2.8cm} p{2.1cm}}
\toprule
\textbf{Category} & \textbf{Model Name} & \textbf{Accuracy (\%)} \\
\midrule
MLP & - & 70.44\\
\midrule
\multirow{6}{*}{Standard GNNs}
 & GAT        & 58.21\\
 & GCN             & 60.54\\
 & GIN             & 63.02 \\
 & GATv2       & 65.14 \\
 & GraphSAGE & 72.17\\
 & PNA          & \textcolor{blue}{\textbf{76.44}} \\
\midrule
\multirow{4}{*}{Baselines}
 & RAG-GAT $^\dagger$ & 45.93 \\
 & DISCO-GCN$^\dagger$   & 70.01\\
 & HGNN$^\dagger$        & 70.61 \\
 & IMGCN-LPE$^\dagger$ & 73.09 \\
 & ShapeGNN & \textcolor{green}{\textbf{80.18}} \\
\midrule
Ours & SuperGT                           & \textcolor{red}{\textbf{80.19}} \\
\bottomrule
\end{tabular}
\end{table}

\subsection{Optimal PE/SE}
\label{sec:optimalPSE}
As discussed in Section~\ref{subsec:gt}, we consider LapPE, ElstaticPE, RWSE and GPSE variants for finding an optimal PE/SE for image classification task using superpixel graphs and graph transformer. For each PE/SE, we experiment with two types of $\text{GlobalAttn}^{\ell}$: (i) Transformer \cite{vaswani2017attention}, and (ii) Performer \cite{choromanskirethinking}. 
Table~\ref{tab:pse_comparison} presents a comparison for the results obtained using these models. We notice a marginal improvement in performance by using PE/SE compared to using none. Interestingly, the similar trend has been observed in previous works on CIFAR-10 KNN superpixel graphs \cite{canturkgraph, rampavsek2022recipe}. The plausible reason for such observation indicates that  $\text{MPNN}^{\ell}$ in the layer might be able to capture the inherent structural information of the superpixel graph.

Nevertheless, we record that the three best performing PE/SE augmentations under the Performer mechanism belong to the GPSE~\cite{canturkgraph} family. This is favorable because the Performer attention mechanism and the GPSE \cite{canturkgraph} scale linearly with the number of nodes. In contrast to Laplacian-based encodings such as LapPE \cite{kreuzer2021rethinking, dwivedi2023benchmarking}, GPSE \cite{canturkgraph} embeddings are outputs of a pre-trained GNN and do not require computationally expensive matrix decompositions. Among all the variants, GPSE \cite{canturkgraph} pretrained on GEOM~\cite{axelrod2022geom} achieves the best performance. Therefore, we used this particular combination in all subsequent experiments using the proposed model, namely, \texttt{SuperGT}.

\subsection{Performance of \texttt{SuperGT} compared to the baselines}
\label{sec:Performance}
We compare \texttt{SuperGT} with popular GNN models such as \texttt{GCN} \cite{kipf2017semi}, \texttt{GAT} \cite{velivckovic2018graph}, \texttt{GATv2} \cite{brodyattentive}, \texttt{GraphSAGE} \cite{hamilton2017inductive}, \texttt{PNA} \cite{corso2020principal} and \texttt{GIN} \cite{xupowerful} along with several segmentation-based GNNs such as \texttt{DISCO-GCN} \cite{youn2021dynamic}, \texttt{RAG-GAT} \cite{avelar2020superpixel}, \texttt{HGNN} \cite{long2021graph}, \texttt{IMGCN-LPE} \cite{bae2022superpixel} and \texttt{ShapeGNN} \cite{cosma2023geometric}. Furthermore, we also consider a simple Multi-Layer Perceptron (\texttt{MLP}). The results are shown in Table~\ref{tab:model_comparison}.

It is evident from Table~\ref{tab:model_comparison} that simple \texttt{MLP} performs better than message passing GNNs such as \texttt{GAT}, \texttt{GCN}, \texttt{GIN}, \texttt{GATv2}, and \texttt{GraphSAGE}. This suggests that local message passing may be insufficient for effectively modeling RAGs, motivating the exploration of more complex graph learning models. However, \texttt{PNA} and \texttt{GraphSAGE} shows considerably better performance than simple MLP.

\texttt{SuperGT} demonstrates a strong performance when compared to the best in standard GNN, i.e., \texttt{PNA} and appropriate baselines for superpixel-based image classification except \texttt{ShapeGNN}. Although \texttt{SuperGT} performs comparable to \texttt{ShapeGNN}, it should be noted that unlike \texttt{ShapeGNN}, \texttt{SuperGT} is invariant to translations of the underlying image and does not require coordinates of the boundary points of each superpixel. The superior performance of \texttt{SuperGT} can be attributed to the ability of $\text{GlobalAttn}^{\ell}$ to capture long-range dependencies.

\begin{figure}[t]
\centering
\begin{subfigure}{.48\linewidth}
    \centering
    \includegraphics[width=\linewidth]{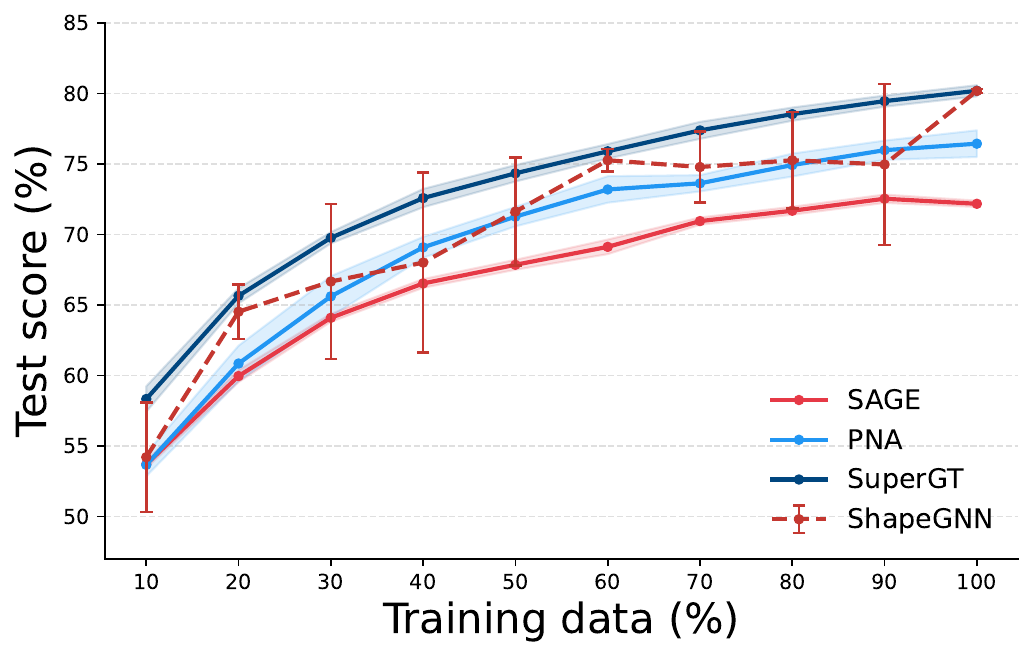}
    \caption{Training on subsets}
    \label{fig:sample-train}
\end{subfigure}
\hfill
\begin{subfigure}{.48\linewidth}
    \centering
    \includegraphics[width=\linewidth]{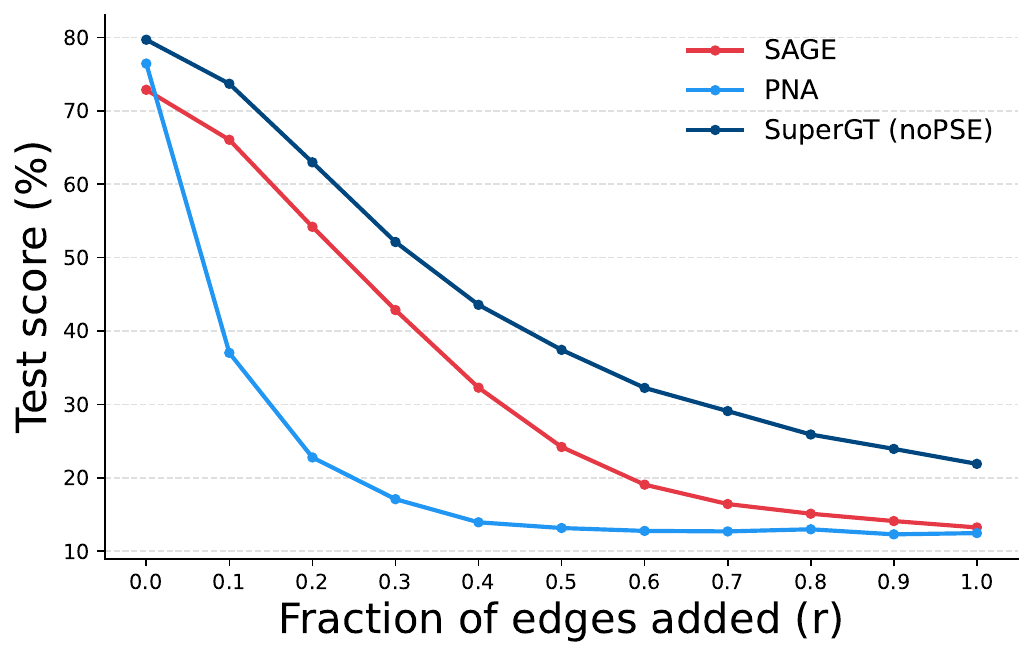}
    \caption{Random Edge Addition}
    \label{fig:edge-add}
\end{subfigure}

\vspace{1em}

\begin{subfigure}{.48\linewidth}
    \centering
    \includegraphics[width=\linewidth]{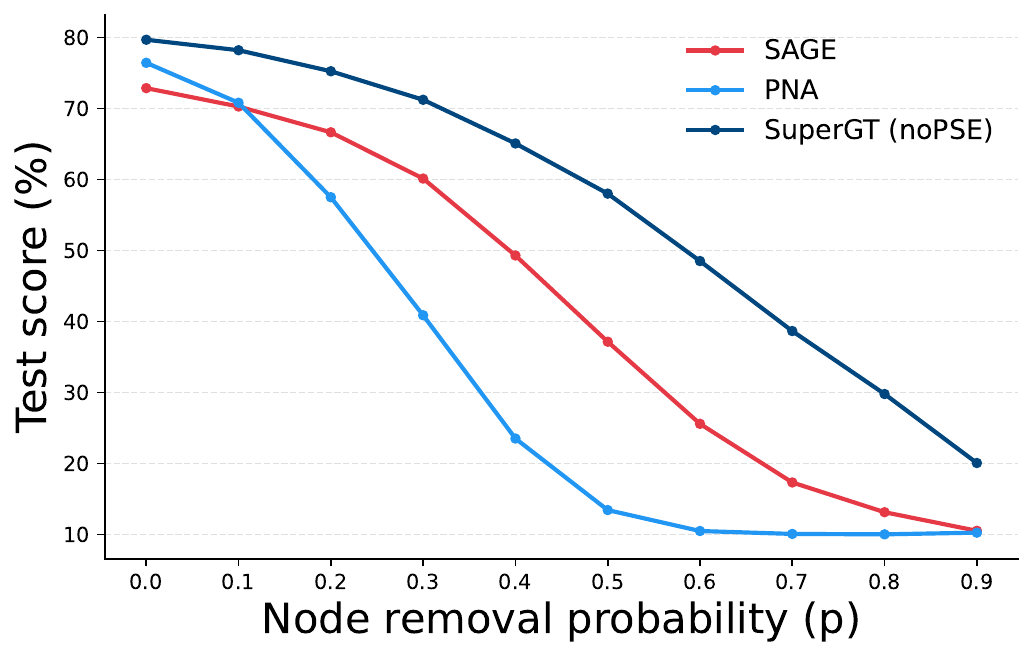}
    \caption{Random Node Removal}
    \label{fig:node-remove}
\end{subfigure}
\hfill
\begin{subfigure}{.48\linewidth}
    \centering
    \includegraphics[width=\linewidth]{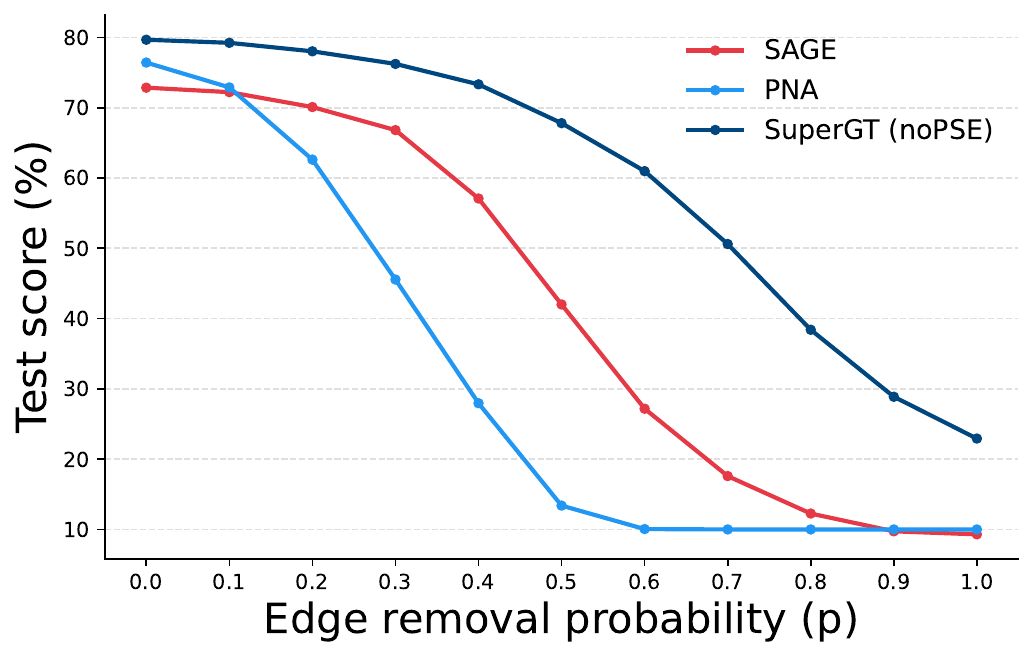}
    \caption{Random Edge Removal}
    \label{fig:edge-remove}
\end{subfigure}
\caption{Robustness analysis.}
\label{fig:robustness}
\end{figure}







\paragraph{\textbf{Performance under limited training data}} Now we evaluate \texttt{SuperGT} on the same test split while training it on different sizes of training samples. For each training set size, we generate different training subsets using 5 random seeds and evaluate the resulting models on the fixed test set, allowing us to estimate the mean and variability of model performance. We perform this experiment for \texttt{ShapeGNN} and two top performing GNN baselines, \texttt{PNA} and \texttt{GraphSAGE}. We show the results in Figure~\ref{fig:sample-train} and it is evident that \texttt{SuperGT} outperforms other models across different sizes of training data. \texttt{ShapeGNN} exhibits noticeably higher performance variance at smaller training set sizes. This indicates high sample efficiency and high robustness of \texttt{SuperGT} to limited supervision. 

\subsection{Robustness to structural noise} We evaluate the robustness of models under three types of structural noise applied to the test graphs: (i) randomly adding edges, where the number of added edges is a fraction $f$ of existing edges, (ii) randomly removing nodes with probability $p$ using samples from a Bernoulli distribution, (iii) randomly removing edges with probability $p$ using samples from a Bernoulli distribution. We show the results for each type in Figures~\ref{fig:edge-add},\ref{fig:node-remove} and \ref{fig:edge-remove} respectively. To isolate the effect of structural perturbations, we omit PE/SE in this experiment.  We observe that for each of these experiments, our model outperforms other models across all values of $p$ and $f$. Furthermore, the performance of \texttt{SuperGT} degrades more gracefully as the noise level increases. This can be attributed to the fact that GNNs rely mainly on the local structural information whereas \texttt{SuperGT} can leverage global information using the $\text{GlobalAttn}^{\ell}$.

\subsection{Ablation Studies}
\label{sec:ablation}
\paragraph{\textbf{Effect of PCA-based pre-processing}}
We proposed a PCA based pre-processing approach in Section~\ref{subsec:pca} to incorporate both rotation and translation invariance with respect to the underlying image. Table~\ref{tab:ablation} demonstrates performance by PCA-based processing. Surprisingly, using this pre-processing leads to decrease in performance by 2.5\% when compared to simple node centering-based pre-processing. However, we note that this approach indeed performs better than multiple existing graph-based baselines in Table~\ref{tab:model_comparison}. 


\paragraph{\textbf{Traditional shape encoding degrades performance}} Instead of using PE/SE, we explore utilizing shape encoding defined by the boundary of each superpixel. Hu moments \cite{hu1962visual} generate a 7 dimensional representation for a given shape that is \textit{translation}, \textit{rotation} and \textit{scale} invariant. To reduce boundary noise introduced by the segmentation algorithm, we simplify superpixel contours using a polygon approximation algorithm, Douglas–Peucker algorithm \cite{ramer1972iterative}, and then compute Hu moments \cite{hu1962visual} to obtain the shape descriptor. As demonstrated by Table~\ref{tab:ablation}, this method led to a marginal performance decrease of 0.3\%. This marginal decrease doesn't allow for a proper assessment of using this shape encoding.

\paragraph{\textbf{Spatial coordinates improve performance}} We ablate the spatial coordinates in the node attributes. This is a simple way to introduce rotation and translation invariance with respect to the underlying image. This led to a drastic decrease in performance of 5.7\%. A probable reason for this is that without the spatial information the model will be unable to differentiate between structurally similar superpixel connectivity that differ only in their scale or shear within the image, which might be critical for resolving visually ambiguous classes. This observation is consistent with the findings of Rodrigues et al.~\cite{rodrigues2024graph}.

\begin{table}[t]
\centering
\caption{\textbf{Ablation Study:} Reported scores are averaged across 5 random seeds.}
\label{tab:ablation}
\begin{tabular}{lc}
\toprule
Ablation Type & Accuracy \\
\midrule
PCA preprocessing     & 77.68 \\
Centroid coordinates  & 74.50 \\
KNN graphs            & 75.39 \\
Hu-moments            & 79.92 \\
Translation only                  & \textbf{80.19} \\
\bottomrule
\end{tabular}
\end{table}

\paragraph{\textbf{KNN graphs degrade performance}} We now evaluate \texttt{SuperGT} on KNN graphs derived from CIFAR-10 dataset following the procedure proposed by Dwivedi et al \cite{dwivedi2023benchmarking}. This led to a drastic decrease in performance of 4.8\% when compared to the proposed \texttt{SuperGT} model where graphs are constructed using RAG. 

\section{Conclusion}
In this work, we proposed \texttt{SuperGT}, a graph transformer-based framework that addresses the long range dependency limitation for superpixel-based image classification. As the traditional image classifiers like CNNs preserve translation invariance, we revisit and propose pre-processing techniques which can preserve translation and rotation invariance in superpixel-based graphs. In particular, we present a Principal Component Analysis (PCA)-based framework capable of preserving both of the invariance simultaneously. We utilize graph transformer to capture long range information and evaluate the proposed framework \texttt{SuperGT} on CIFAR-10 benchmark data. With suitable experiments, we show that \texttt{SuperGT} achieves state-of-the-art performance (comparable to previous state-of-the-art with marginal improvement) when models are trained on all the training samples. We demonstrate that \texttt{SuperGT} demonstrates stronger performance than state-of-the-art when subjected to different variance of training samples. Thus, we conclude that Graph Transformers can serve as an effective alternative to message-passing GNNs for superpixel-based image classification while incorporating desirable geometric invariance properties.






\bibliographystyle{elsarticle-num}
\bibliography{references}






\end{document}